\documentclass[review]{elsarticle}
\usepackage{fix-cm}
\usepackage{mathrsfs}

\usepackage{tabularx}  
\usepackage{amssymb}
\usepackage{multirow}
\usepackage{multicol} 
\usepackage{array}
\newcolumntype{C}[1]{>{\centering\arraybackslash}p{#1}}
\usepackage{caption} 
\usepackage{graphicx}
\usepackage{mathrsfs}
\usepackage{subfigure}      
\usepackage{xcolor}
\usepackage{hyphenat}
\usepackage{amsmath}
\usepackage{subcaption}
\usepackage{subfigure}
\usepackage{float}
\usepackage{amsmath,amssymb,amsfonts}%
\usepackage{amsthm}%
\usepackage{mathrsfs}%

\makeatletter
\def\ps@pprintTitle{%
  \let\@oddhead\@empty
  \let\@evenhead\@empty
  \def\@oddfoot{}%
  \let\@evenfoot\@oddfoot
}
\makeatother

\let\mathscr\mathcal
\begin{document}

\begin{frontmatter}



\author[label1]{Mobina Mobaraki}
\author[label2]{Mahyar Asasdi}
 \author[label3] {Klaske Van Heusden}
 \author[label1] {Guy A. Dumont}

\affiliation[label1]{organization={University of British Columbia- Vancouver campus},
             addressline={2329 West Mall},
             city={Vancouver},
             postcode={V6T 1Z4},
             state={British Columbia},
             country={Canada}}

 \affiliation[label2]{organization={Novarc Technologies Inc.},
             addressline={1225 E Keith Rd},
             city={North Vancouver},
             postcode={V7J 1J3 },
             state={British Columbia},
             country={Canada}}

\affiliation[label3]{organization={University of British Columbia, Kelowna campus},
             addressline={3333 University Way},
             city={Kelowna},
             postcode={V1V 1V7},
             state={British Columbia},
             country={Canada}}

\title{Explainable Temporal Attention-based Defect Detection For Fillet Joints in Real-Time Gas Metal Arc Welding Based on Multi-modal Data}

             








\begin{abstract}
 {Deep learning is an efficient technique to monitor the real-time welding process, reducing post-welding repairs and production delays. This paper leverages the monitoring capability by proposing a multi-modal temporal attention-based deep learning defect detection model for internal defects that are challenging to detect, including porosity, lack of penetration and fusion, undercut, and cold lap during Gas Metal Arc Welding in fillet joints. The model is trained on collected welding images and sound data from an industrial collaborative welding robot. The results show that the attention module can improve the F1 Score to 0.99. We use explainable Artificial Intelligence to interpret the proposed models' behavior and dataset distribution, determining potential important areas in image and sound spectrograms and preferred modality to detect each defect. This improves trust and reliability in Artificial Intelligence-driven welding inspection.} 
\end{abstract}



\end{frontmatter}



\section{Introduction}
\label{sec:introduction}
Real-time defect detection is necessary to improve the efficiency of robotic welding~\cite{intro67, nneww1}.
To detect defects, some studies used design feature-based models~\cite{intro24} 
to manually extract geometrical features of the molten pool. These approaches extract predefined, low\hyp{}dimensional image features and struggle to adapt to the differences in the welding torch structure!\cite{nneww2} and dynamic changes of arc morphology in Gas Metal Arc Welding (GMAW)~\cite{newnew1}. Especially, fillet welds in GMAW-S (short-circuiting) mode present rapidly changing conditions due to high-frequency arc cycles and quick-freezing weld puddles, challenging the design feature-based methods and necessitating more adaptive techniques.~\cite{newnew2}. 



More robust alternatives are deep learning-based methods including Convolutional Neural Networks (CNNs)~\cite{newintro3}  
that progressively extract high\hyp{}dimensional features from the raw data through multiple convolutional layers. 
However, they face challenges in real-time processing of welding data~\cite{newintro4, newintro36, intro39} due to their limited ability to store long-term spatiotemporal information. Loss of information from previous time steps can reduce the defect detection performance.

As a solution, an attention mechanism can be integrated into CNNs. Attention mechanisms are inspired by the human nervous system to prioritize salient information. They enhance the efficiency of CNNs by assigning attention weights and prioritizing the previous temporal information~\cite{newintro5}. These mechanisms determine the matching between the extracted features and task-specific representation vector and convert them to attention weights through a distribution function (Softmax)~\cite{newintro6}. Channel attention is a subset of attention technique that applies attention weights on the channels of data sequence~\cite{newintro10}. In welding applications, channel attention focuses on critical spatial information from weld pool images~\cite{newintro7}, such as pool shape, size, and surface patterns, as well as temporal information from acoustic spectrograms~\cite{newintro10, newintro35}. This incorporates subtle changes in weld pool dynamics and acoustic signatures over a longer period of time, enhancing defect detection performance compared to conventional CNNs without attention.

Channel attention can be used to emphasize the important welding features with respect to time, frequency, and modality.
Firstly, the temporal features  
are vital to detect the real-time hardly-detectable internal defects such as porosity~\cite{newintro4} and partial penetration~\cite{newintro11}. As an example, a previous study~\cite{newintro11} found the correlation between variations in arc voltage and molten metal's geometry to determine the level of penetration. Channel attention can reduce the side effects of arc disturbances by focusing on the important time steps. Secondly, the attention mechanism can emphasize the interesting regions in the frequency band of the arc sound signal based on the characteristics of each defect ~\cite{newintro12}. Thirdly, 
fusing the data from multiple sources can provide larger amount of information~\cite{nneww4, newintroo}. 
Channel attention can elaborate on information from sources with less noise interference and redundancy.




Failure to scientifically interpret the defect detection models results in an inability to fix the issue in case of failure~\cite{newintro20}, lack of trust by the users~\cite{newintro14}, and consequently limiting the application in safety-sensitive tasks such as welding~\cite{newintro1}. Explainable Artificial Intelligence (XAI) is a solution to make a deep learning model's decision more transparent. In case of not-reliably deployable models, XAI identifies failures and enable researchers to focus on low-confidence cases that may indicate small defects. In case of reliably deployable ones, XAI improves trust.  

XAI techniques  are divided into Global (GX) and Local (LX) Explainability, with the former focusing on model's learning and the latter focusing on model's assessment. CAM~\cite{newintro19} methods as subsets of LX, highlight the important regions on the images by using backpropagation gradient and feature map of the convolutional layer~\cite{newintro18}. For the Classic CAM (C-CAM) approach, users need to make structural changes on the deployed models.

Advanced CAM techniques such as GradCAM~\cite{newintro20} and GradCAM++~\cite{newintro21} have been proposed to explain the model without altering the structure. Grad CAM calculates the weights for the explanation map by back-propagating the predictions of the last fully connected layer. Unlike other CAMs, Grad CAM is applicable to CNNs with multi-modal inputs~\cite{newintro20}. GradCAM++ is an enhanced version of GradCAM that determines the weights as the predetermined class scores. 

In welding applications, Zhao et al.~\cite{newintro12} used Grad CAM to explain their defect detection model in laser welding by determining the areas of their model's interest in acoustic data. With 98.25 \% accuracy, they claimed that lack of penetration is associated with higher frequency range (15 - 20 kHz), while normal penetration and burn through are linked to mid-range (6-10 kHz) and low-range (2 - 6 kHz) frequencies, respectively. Liu et al.~\cite{newintro1} also used CAM based on Multi-Scale Fusion Features (CAM- MSFF) to provide an accurate (98.76 \%) and human-comprehensible defect detection than the conventional CNNs in TIG welding. They analyzed the distribution of their data using t-SNE. Despite the limited studies to explain defect detection in Tungsten Inert Gas (TIG) and laser welding, to the best of our knowledge, there is no study to interpret defect detection in GMAW. GMAW has more unstable arc with more obscuring elements (consumable electrode, spatter, and smoke) compared to laser welding and TIG~\cite{newintro_}. This paper aims to fill this gap using GradCAM, GradCAM++, and XGradCAM to interpret the defect detection models as well as t-SNE and manual feature analysis to interpret the welding dataset in GMAW . 

In our recent study, we compared the performance of the vision-based, sound-based, and multi-modal CNNs to detect five target defects including lack of penetration (LOP), lack of fusion (LOF), porosity (Por), undercut (U), and cold lap (C) in fillet joints during a real-time GMAW process showing that multi-modal defect detection improves the F1 Score to classify the mentioned defects
by 3.53 \%, 58.62 \%, 30.14 \%, 11.76 \%, and 46.67 \%, respectively. In this work, we first propose a temporal multi-modal defect detection model. Second, we apply an attention technique to focus on the important temporal and frequency features to improve the defect detection performance and compare the results with the normal multi-modal results from our previous work. Third, we analyze the spatial and temporal-frequency behavior of our proposed defect detection models in GMAW in order to understand the important features to detect each target defect. We use Class Activation Map (CAM)~\cite{newintro19} techniques to interpret our models behavior as well as T-distributed stochastic neighbor embedding (t-SNE)~\cite{tsne} and manual feature analysis to interpret the distribution of our welding dataset. The summary of our main contributions are as follows.

\begin{itemize}  
    \item \textbf{Enhanced defect detection by adding temporal features: } For a more accurate and robust defect detection, we consider the effect of time in the data and improve the F1 Score by 7.95 \%, 6.52 \%, 4.21 \%, 4.21 \%, and 12.5 \% to detect LOP, LOF, porosity, undercut, and cold lap, respectively.

    \item \textbf{Enhanced defect detection using attention techniques: } To improve the performance of defect detection models, we focus on the most useful data within a period of time and frequency and improve the performance of the defect detection model by 4.21 \% and 1.02 \% for LOP and LOF, respectively, leading to an F1 Score of 0.99 for all the five target defects with a minimum additional computational cost of 0.1 GFLOPs.

    \item \textbf{Interpretability of defect detection models: } We explain our proposed defect detection models with XAI techniques including GradCAM, GradCAM++, and XGradCAM and determine the speculative important areas in image and sound data to detect each defect.

    \item \textbf{Interpretability of defect detection dataset: } We explain our dataset with t-SNE and manual feature analysis and determine the preferred modality to detect each defect.
\end{itemize}

\section{Equipment and dataset}
\label{sec:equ}


\subsection{Equipment}

\begin{figure}[H]\centering
	\includegraphics[width=0.8\textwidth]{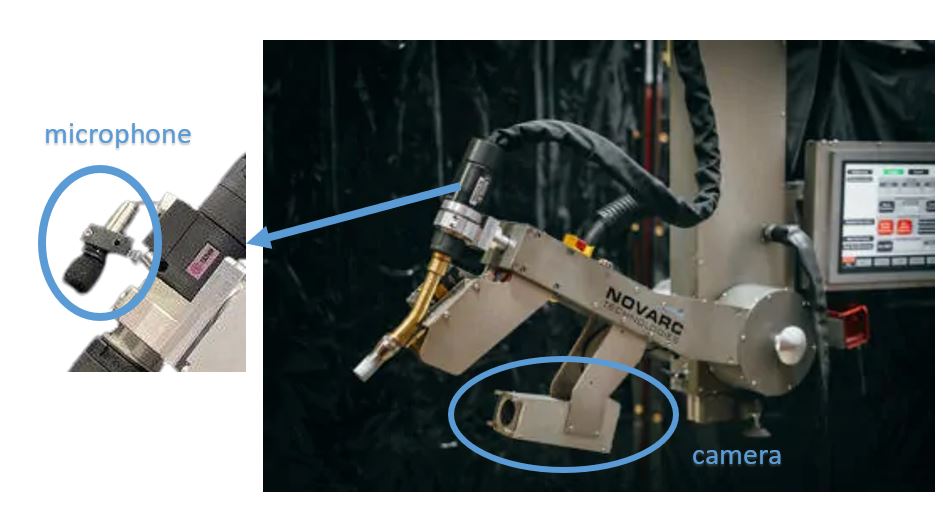}
	\caption{The industrial collaborative welding robot used to capture the welding image and sound data}\label{fig:swr} 
\end{figure}

The experimental setup involved a collaborative welding robot, illustrated in Figure~\ref{fig:swr}, which was employed to capture welding images and sound data. The robot is equipped with three servo motors, enabling precise torch movement in seam, weave, and distance directions. The power source was a Miller Auto-Continuum™ 500, with 85/15 welding wire and a shielding gas mixture of Argon and CO2, at a flow rate of 35–55 cubic feet per hour (CFH). A Complementary Metal Oxide Semiconductor (CMOS) camera from (NovEye\textsuperscript{TM} at Novarc Technologies Inc.) was mounted on the robot's torch, maintaining a distance of 20–25 cm from the molten metal. This camera captured video footage of the molten metal in the fillet joints during the Gas Metal Arc Welding (GMAW) process, with a sampling frequency of 16 frames per second (fps) and a resolution of $(1680\times1240\times3)$. To mitigate the brightness of the welding arc, a filter was affixed to the camera lens. Additionally, a microphone, also attached to the torch (Figure~\ref{fig:swr}), recorded the welding sound at a sampling frequency of 51,200 Hz. The tests were conducted in a research and development (R\&D) environment, characterized by lower noise levels compared to a typical production setting.

\subsection{Data}

We collected images and sound data through 8 welding tests on fillet joints between a carbon steel pipe with 6” diameter and a flange during the GMAW process. For each test, the pipe was divided into six sections, and the welder deliberately created either defective or defect-free welds in each section. The targeted defects included lack of penetration (LOP), lack of fusion (LOF), undercut, cold lap, and porosity. Table~\ref{tab:testplan} outlines the test plan for the eight welding tests. Table~\ref{tab:testinfo} details the procedures for generating and labeling each targeted defect. Table~\ref{tab:testpar} presents the welding parameters and their corresponding values used during the tests.

\begin{table}[H]
    \renewcommand{\arraystretch}{1.3}
    \centering
    \centering
    \resizebox{\textwidth}{!}{ 
        \begin{tabular}{lclclclclclclcl}
            \hline\hline \\[-3mm]
            Pipe \# & Type & Section 1 & Section 2 & Section 3 & Section 4 & Section 5 & Section 6\\
            \hline\hline \\[-3mm]
            \multirow{2}{4em}{Pipe 1} & Inner & Normal & LOP & Normal &  LOP & Normal & LOP \\
            & Outer & Normal & LOP & Normal &  LOP & Normal & LOP \\
            
            \multirow{2}{4em}{Pipe 2} & Inner & Normal & Porosity & Normal &  Porosity & Normal & Porosity \\
            & Outer & Normal & Porosity & Normal &  Porosity & Normal & Porosity \\

            \multirow{2}{4em}{Pipe 3} & Inner & Normal & LOF & Normal &  Cold Lap & Normal & LOF \\
            & Outer & Normal & LOF & Normal &  Cold Lap & Normal & LOF \\

            \multirow{2}{4em}{Pipe 4} & Inner & Normal & Undercut & Normal &  Undercut & Normal & Undercut \\
            & Outer & Normal & Undercut & Normal &  Undercut & Normal & Undercut 
            \\
            \hline\hline
        \end{tabular}
    }
    \caption{Test plan to collect the representative image and sound data as training data for the deep learning defect detection models.}\label{tab:testplan}
\end{table}

\begin{table}[H]
	\renewcommand{\arraystretch}{1.3}
	\centering
    \centering
	\resizebox{\textwidth}{!}{
		\begin{tabular}{lclcl}
                \hline\hline \\[-3mm]
                Defect & How to create & Post welding test\\
			\hline\hline \\[-3mm]
               LOP & Reduce current & \hfil Ultrasound\\
               LOF & Reduce current, Excessive heat & \hfil Ultrasound\\
               Undercut & Insufficient weave amplitude,
Excessive heat & \hfil Visual inspection\\
               Cold Lap & Stop weaving, Excessive heat& \hfil Visual inspection\\
               Porosity & Grease, Fan & \hfil Visual inspection\\
               \hline\hline
		\end{tabular}
	}
    \caption{The five target defects in our welding tests and methods to create the defect during the welding process and detect them in the post welding evaluation}\label{tab:testinfo}
\end{table}

\begin{table}[H]
    \renewcommand{\arraystretch}{1.3}
    \centering
    \resizebox{0.6\textwidth}{!}{
    \begin{tabular}{C{5cm} C{4cm}}
        \hline\hline \\[-3mm]
        Welding Parameters & Values \\[1.6ex] \hline
        \hline
        Average current (A) & 70 - 200 \\
        Voltage (V) & 17 - 35\\
        Wire Feed Speed (IPM) & 100 - 240\\
        Travel speed (IPM) & 9.5 - 17.5\\
        \hline\hline
    \end{tabular}
    }
     \caption{Welding parameters range during data capturing}\label{tab:testpar}
\end{table}

The size of training, validation, and test datasets are 23543, 2942, and 2942 images, respectively. The dataset in this study is larger and more representative than the average of 500-3000 images in other studies~\cite{intro65} which makes our proposed defect detection models more robust in different welding scenarios. Table~\ref{tab:datanum} shows the number of video frames for each class in our dataset.  
A video annotation interface~\cite{new20} was used to label the images. 

\begin{table}[H]
	\renewcommand{\arraystretch}{1.3}
	\centering
        \resizebox{0.4\textwidth}{!}{
		\begin{tabular}{C{3cm} C{3cm} C{3cm}}
                \hline\hline \\[-3mm]
                Class & \# images \\
			[1.6ex]\hline
                \hline
               LOP & 3290 \\
               LOF & 4131 \\
               undercut & 1385 \\
               cold lap & 2466 \\
               porosity & 5492 \\ 
               defect-free & 12665 \\
			\hline\hline
		\end{tabular}
  }
  \caption{Total number of welding images for each class in our dataset}\label{tab:datanum} 
\end{table}

Sound signals were transformed into mel spectrograms~\cite{nneww3}. This representation is widely used in sound event classification~\cite{new19} and includes both temporal and frequency-domain features of the sound signals~\cite{new17}. It also enables the human-like perception by focusing on the lower frequencies and producing a more informative sound images. To generate mel spectrogram, a window size of 3200 is selected to collect the same number of image and sound spectrograms per second. This is the sound’s sampling frequency (51200 Hz) over image’s sampling frequency (16 fps). Also, a minimum frequency of 20 Hz, a maximum frequency of 20,000 Hz, a stride size (HOP) of 2500, and 10 mel bands were selected~\cite{equipment4,intro14}.


\section{Methods}
\label{sec:model}
\subsection{Multi-modal models}

In our previous study, ResNet 18 2D~\cite{equipment1} model was used for the image-based and sound-based uni-modal defect classification models. To train the multi-modal model, sound and image feature vectors from the uni-modal models were fused from different layers of the uni-modal models and different fusion techniques including concatenation, adding, linear regression, and contrastive learning methods. The results from our previous work are shown in Table~\ref{tab:multi_fusion}. The Table reports the fusion method that results in the highest F1 Score and lowest inference time to classify each defect.

\begin{table}[H]
	\renewcommand{\arraystretch}{1.3}
	\centering
    \centering
	\resizebox{\textwidth}{!}{
		\begin{tabular}{l l l l l}
            \hline
            \hline
            \hfil Defect & \hfil Best fusion Method & \hfil Feature Layer & \hfil F1 Score & \hfil Time to classify fused feature vector (ms) \\
            
			\hline\hline \\[-3mm]
            \hfil LOP & \hfil Concatenation & \hfil Layer 4 & \hfil 0.88 &\hfil 0.70\\
            \hfil LOF & \hfil Concatenation & \hfil Layer 4 & \hfil 0.92 & \hfil 0.80\\
            \hfil Undercut & \hfil Contrastive learning & \hfil Layer 1 & \hfil 0.95 & \hfil 0.12\\
             \hfil Cold Lap & \hfil Contrastive learning & \hfil Layer 1 & \hfil 0.95 & \hfil 0.40\\
            \hfil Porosity & \hfil Concatenation & \hfil Layer 4 & \hfil 0.88 & \hfil 0.70\\
            \hline\hline
		\end{tabular}
	}
    \caption{The best fusion techniques for each defect detection model to get the highest F1 Score}\label{tab:multi_fusion}
\end{table}

In this study, we conducted an ablation study in Table~\ref{tab:input} to find the input size. The Table compares the F1 score and complexity of the ResNet 18 model (in terms of Floating Point operations per second (GFLOPs)~\cite{flops,flops2}) to detect LOP with different input sizes. We chose $(112\times112)$ as the height and width of the image and spectrogram as further increasing the input size will increase the computational complexity more than 1 GFLOPs and the training becomes 5 times longer. In the next sections we add an attention module to the models that improves the F1 score to 0.99 with an additional computational cost of only 0.1 GFLOPs.

\begin{table}[H]
	\renewcommand{\arraystretch}{1.3}
	\centering
    \centering
	\resizebox{0.6\textwidth}{!}{
		\begin{tabular}{l l l l l l}
            \hline
            \hline
            \hfil Input size & \hfil F1 score & \hfil GFLOPs (\(10^9\)) & \hfil Training time (min)\\
            
			\hline\hline \\[-3mm]
            $(56\times56)$\hfil  & 0.65\hfil  & 0.13 \hfil& 32 \hfil  \\
            $(112\times112)$\hfil  & 0.71\hfil  & 0.48 \hfil & 63 \hfil  \\
            $(224\times224)$\hfil  & 0.88\hfil  & 1.82 \hfil & 305 \hfil  \\
            \hline\hline
		\end{tabular}
	}
    \caption{The input image size to detect LOP based on F1 Score and computational complexity in terms of Floating Point Operations per second (GFLOPs)}\label{tab:input}
\end{table}

\subsection{Temporal Models}

\begin{figure}[H]
    \centering
    \includegraphics[width=0.9\textwidth]{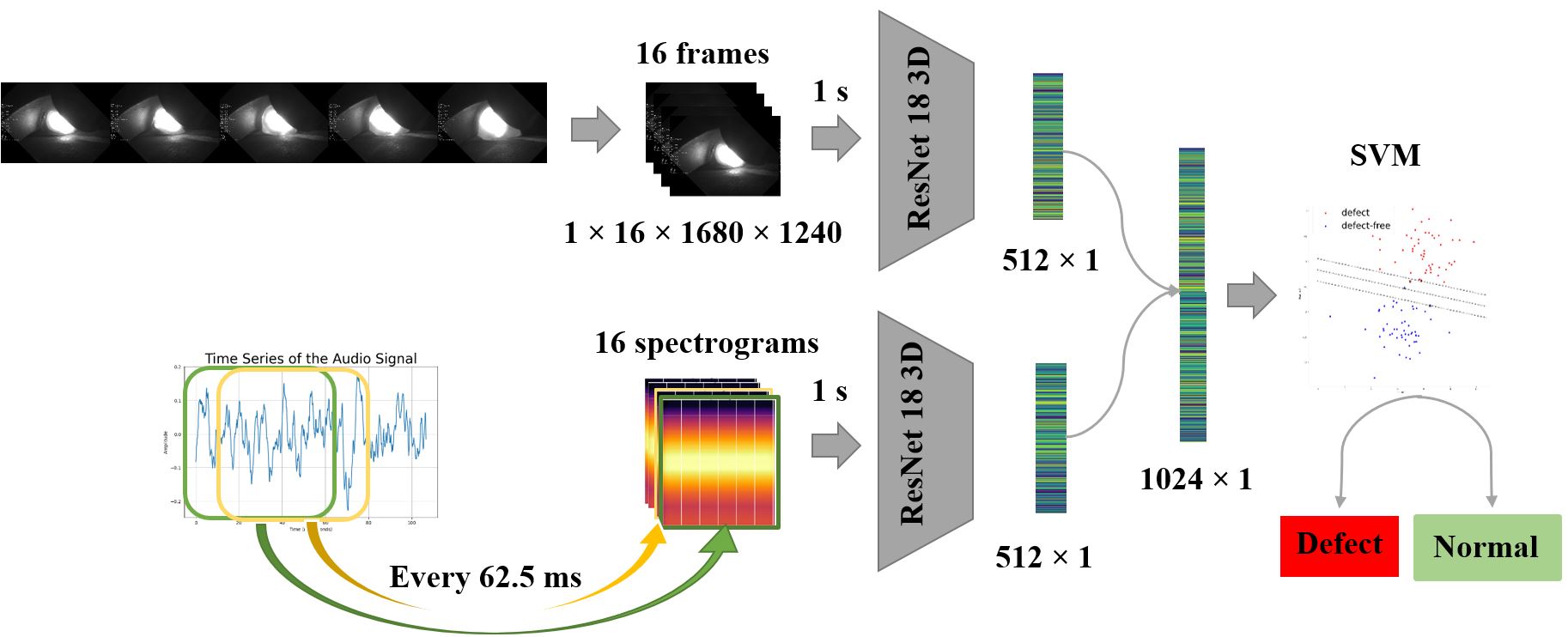}
      \caption{Architecture of the 3D temporal multi-modal defect detection model}
    \label{fig:second}
\end{figure}
\begin{figure}[H]
    \centering
    \begin{multicols}{2} 
        \begin{subfigure}{}
            \includegraphics[width=\linewidth]{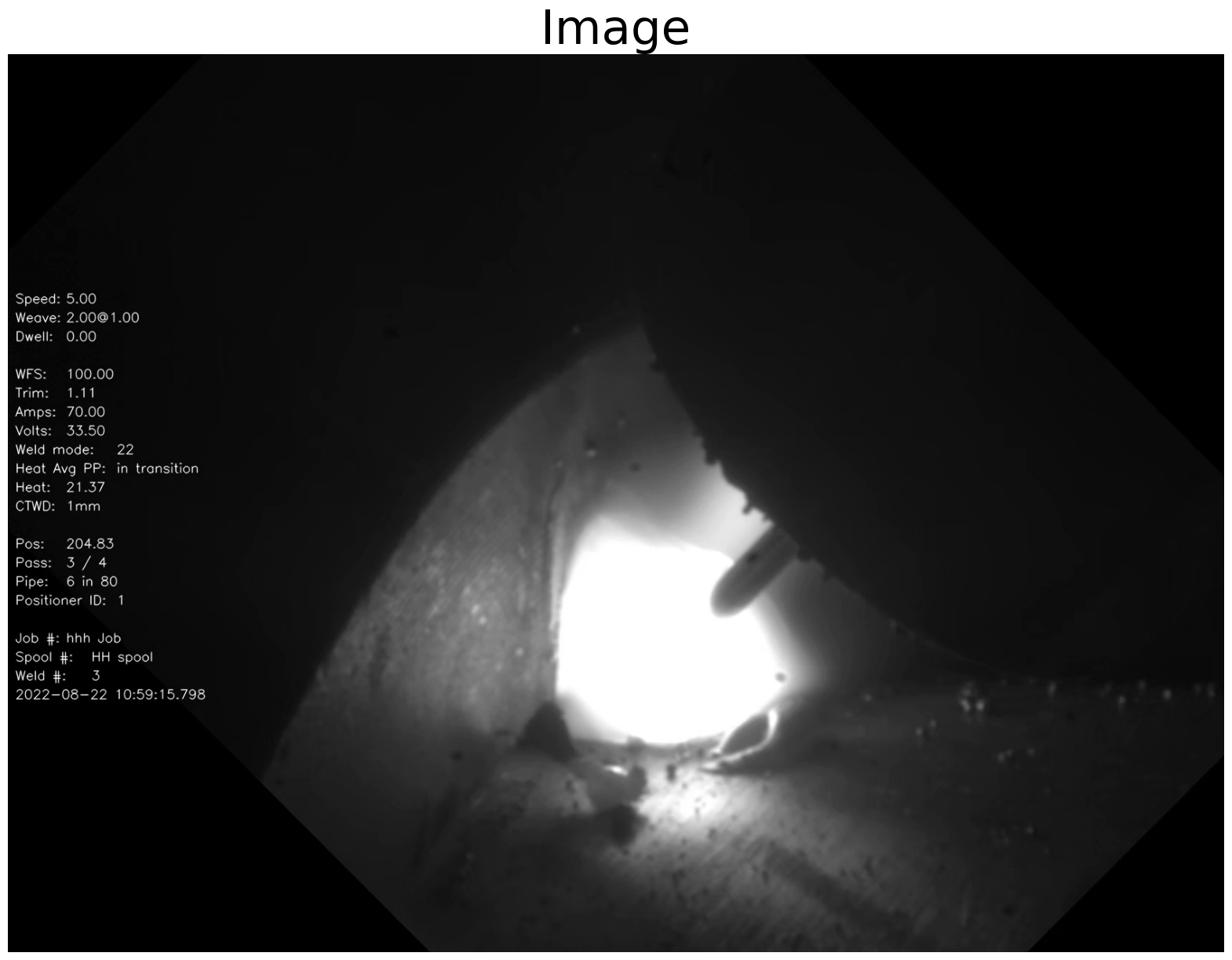}
            \captionsetup{justification=centering}
      
        \end{subfigure}
        \hfill
        \begin{subfigure}{}
            \includegraphics[width=\linewidth]{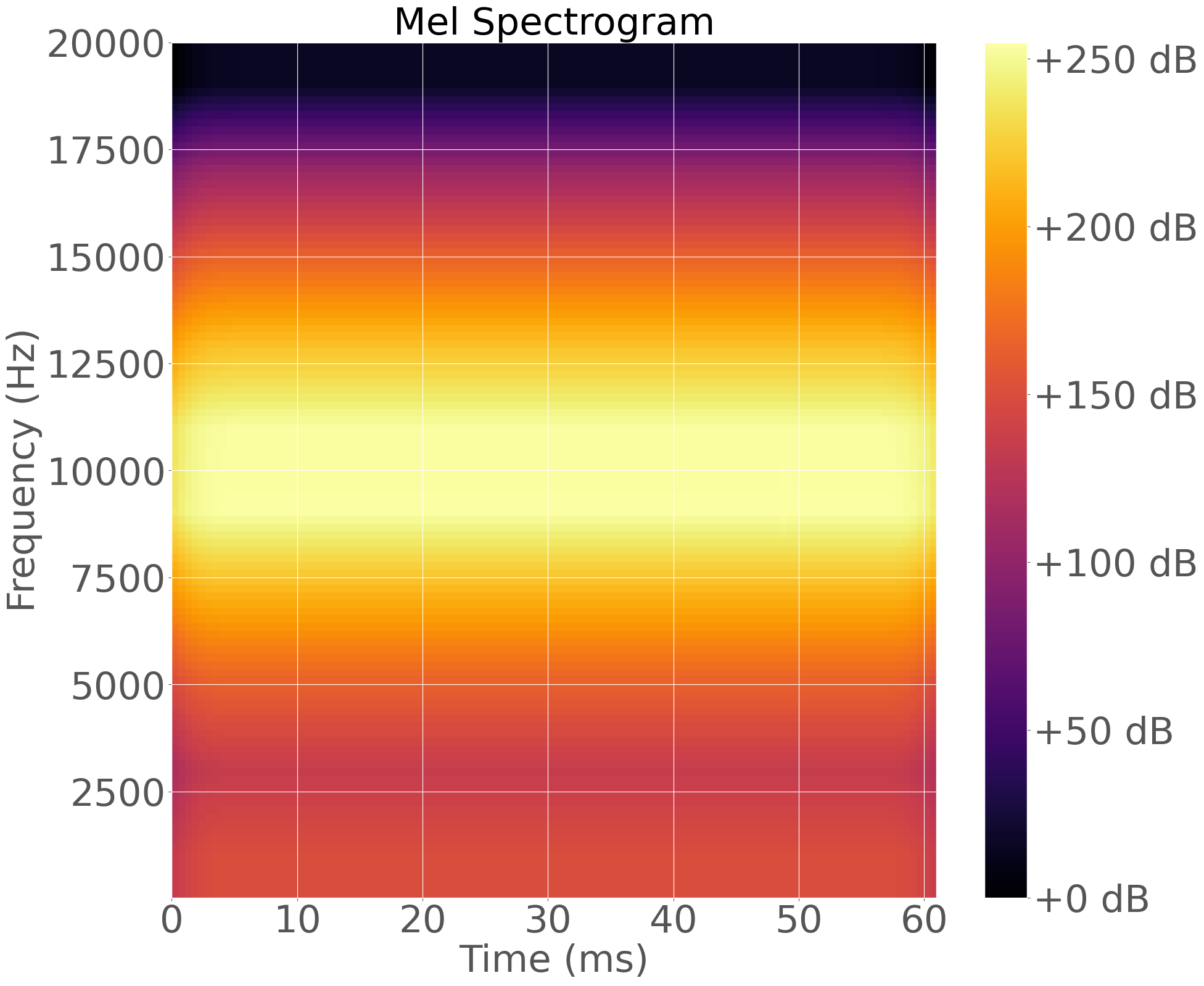}
            \captionsetup{justification=centering}
          
        \end{subfigure}
    \end{multicols} 
    \caption{Samples of image (left) and sound spectrogram (right) as inputs to the temporal ResNet 3D model}
    
    \label{fig:sample}
\end{figure}

We trained a 3D multi-modal model based on ResNet 3D~\cite{newintro27} to include the temporal features as shown in Figure~\ref{fig:second}. In this case, we consider the temporal information as the channel (depth) of the image. The format of the inputs to this model is (batch size, number of channels, number of frames, height, width). Image and sound data have the size of $(16\times1\times16\times1680\times1240)$ and $(16\times13\times16\times10\times103)$, respectively. Figure~\ref{fig:sample} shows a sample of image and sound data.

We stacked 16 images and sound data as a block for every second and removed the last fully connected layer of the ResNet 3D model and applied it on the uni-modal blocks to extract the feature vector of size $(1\times512)$ for each modality.

\subsection{Attention-based temporal models}
\begin{figure}[H]
    \centering
    \includegraphics[width=\textwidth]{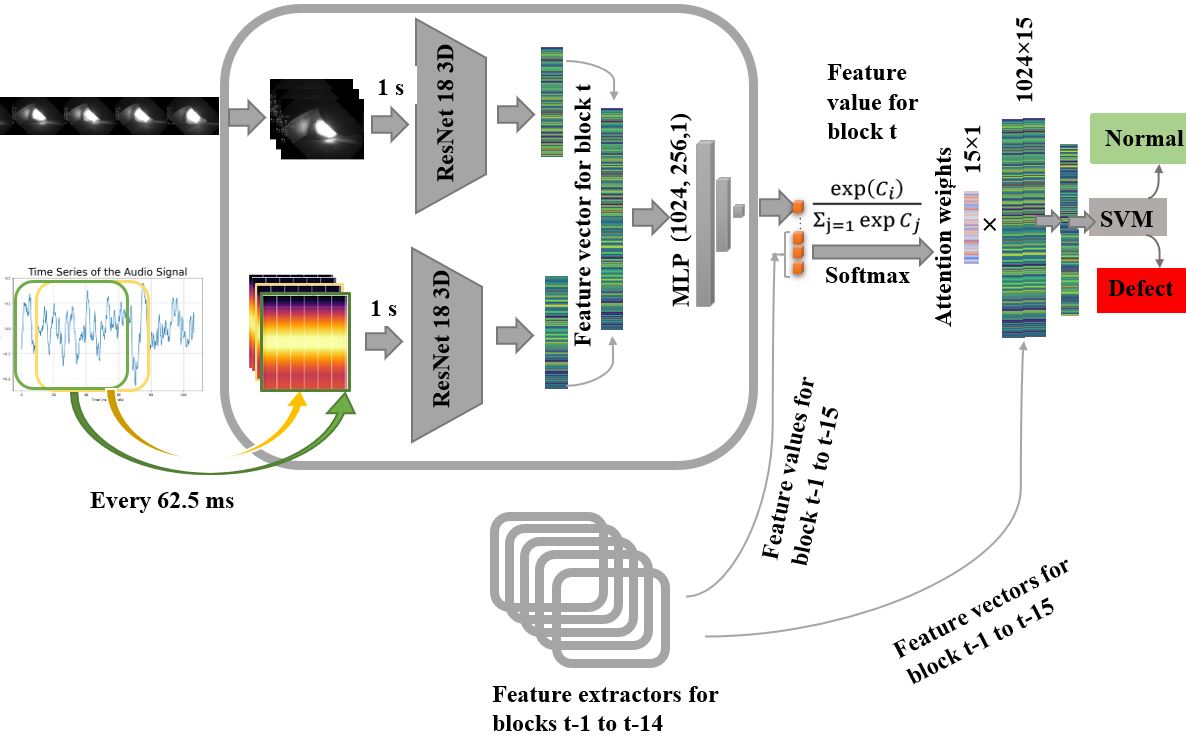}
      \caption{Architecture of the attention-based 3D temporal multi-modal defect detection model}
    \label{fig:third}
\end{figure}

We aimed to improve the temporal multi-modal defect detection models speculating by adding an attention technique to the 3D temporal multi-modal model as shown in Figure~\ref{fig:third}. The proposed model is inspired by the video vision transformer architecture~\cite{new4}. The FLOPs for this approach is 48 \% less than Squeeze-and-Excitation methods~\cite{metrics}.  Also, it is applied to the fused feature maps to consider the cross-model interactions between image and sound data, while SE is applied on the feature maps of each modality separately before the fusion happens~\cite{flops2}.


The data was split into \(n\) blocks of consecutive images and sound spectrograms. Let \(F_{\text{concat},i}\) represents the multi-modal feature vectors that concatenates image and sound feature vectors at block \(i\) of consecutive images and sound spectrograms.

   \[
   F_{\text{concat},i} = [F_{\text{image},i}, F_{\text{audio},i}], \quad i \in \{1, 2, \ldots, n\}.
   \]

We applied a Multi Layer Perceptron (MLP) with 3 layers on the concatenated multi-modal feature vector. This results in a confidence score for each multi-modal block.
   \[
   C_i = \text{MLP}(F_{\text{concat},i}),
   \]
   where \(C_i\) is the confidence score for block \(i\).

The confidence scores were normalized using the Softmax function to compute attention weights, which shows the importance of each block.
   \[
   \alpha_i = \frac{\exp(C_i)}{\sum_{j=1}^{n} \exp(C_j)}, \quad i \in \{1, 2, \ldots, n\}.
   \]

The final weighted multi-modal feature vector was computed by multiplying each multi-modal block by its weight and adding them together.

   \[
   F_{\text{final}} = \sum_{i=1}^{n} \alpha_i \cdot F_{\text{concat},i}.
   \]

A Support Vector Machine (SVM) classifier was trained to find the defective/ defect-free labels from the attention-based multi-modal feature vector.
   \[
   \text{Label} = \text{SVM}(F_{\text{final}}).
   \]

\subsection{Model explainability methods}


\subsubsection{CAM-based XAI techniques}
GradCAM~\cite{newintro14} back-propagates the class scores with respect to the feature map from the last convolutional layer. It calculates weights through Global Average Pooling and applies ReLU activation to only highlight the important pixels of the image with positive weights. 

In this study, we used the following metrics~\cite{newintro21} to evaluate the GradCam results on our proposed defect detection models by perturbing the original image and predicting the results based on the perturbed image.

\begin{itemize}
    \item \textbf{Drop in confidence} calculates the percentage drop of the confidence if we block the unnecessary pixels on the original image based on the explanation map. The confidence is assumed to drop a bit since it removes details. It is a number between 0 and 1 with the smaller values representing the better explainability. 0 value shows that the confidence increased. 

    \item \textbf{Increase in confidence} calculates the number of cases that the confidence is increased by only focusing on the important pixels proposed by the explanation map. Larger values shows that removing the unnecessary details is helpful to improve the model's performance.

    \item \textbf{Percentage metric} removes a percentage (5, 25, and 50 \%) of important pixels proposed by the explanation metric and predicts the output on the remaining pixels. Negative values confirm the importance of the proposed pixels and their effect on model's performance. 
\end{itemize}

For each defect, we first collected the heatmaps of the video frames with positive confidence increase in activation score and 0 value for drop in confidence. The former picks the heatmaps that focus on the suggested speculative parts of the image with higher confidence. The later makes sure to pick the heatmaps with all the necessary pixels on the welding images to detect each defect.

Then, for each defect, we considered three sets of heatmaps with percentage metric of 5 \%, 25 \%, and 50 \% as negative, which means that removing the important area with the mentioned percentage reduces the confidence. This shows the areas with the highest importance to detect each defect. We analyzed the heatmaps of GradCAM on the three sets of heatmaps for each defect.

\subsubsection{F1 score results from the CNN models} This shows the ability of a deep learning model to extract the required information from multi-modal model.

\subsection{Data explainability methods}

\subsubsection{t-SNE }In order to understand the overlap between the classes in the image, sound, and multi-modal data, we performed t-SNE. t-SNE creates a probability distribution that represents similarities between neighbors. It finds the similarity between two data points, which is the conditional probability that the first data point would pick the second one as its neighbor. It has to be proportional to probability density under a Gaussian centered at the first data point.  A perplexity in t-SNE is a target number of neighbors for our central point, with higher values resulting in higher variance~\cite{tsne}. We considered the perplexity as 50.

One of t-SNE evaluation metrics is the Silhouette score~\cite{newintro30}. It is a number between -1 and 1 that determines how similar an object is to its own cluster compared to other clusters. This is a useful metric when the data is imbalanced or the boundaries between clusters are not well defined.

\subsubsection{Manual feature analysis } 


The mean (\(\mu\)) of the images for each defect class was computed pixel-wise using the following formula. Let each image in a class \(i\) be represented as \(I_k \in \mathbb{R}^{H \times W}\), where \(H\) and \(W\) are the height and width of the image, and there are \(N_i\) images in the class.

\[
\mu_i(x, y) = \frac{1}{N_i} \sum_{k=1}^{N_i} I_k(x, y)
\]
where \(I_k(x, y)\) is the pixel intensity at position \((x, y)\) in the \(k\)-th image of class \(i\).

\begin{figure}[H]
\centering
    \centering
    \includegraphics[width=0.3\textwidth]{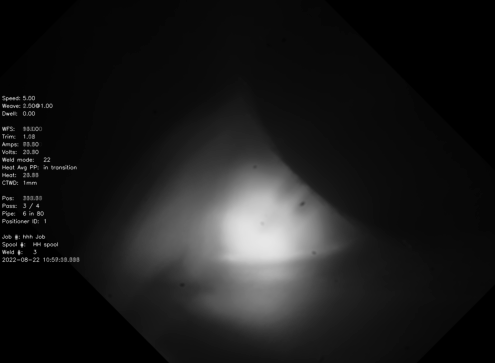}
    
\caption{Visualization of the mean of images for lack of penetration class}
\label{fig:lop_mean_var}
\end{figure}

Figure~\ref{fig:lop_mean_var} shows an example of mean images for lack of penetration.

In this study, we consider two features for class \(i\). 1) The variation, which is the standard deviation of the images in class \(i\) and 2) The contrast between classes \(i\) and \(j\), which is defined as the pixel-wise difference of their mean images:

\[
\text{Contrast}_{i,j}(x, y) = \mu_i(x, y) - \mu_j(x, y)
\]






\section{Results and Discussion}
\label{sec:result}





\subsection{Attention-based temporal defect detection results}
Table~\ref{tab:final} compares the normal, temporal and temporal attention-based multi-modal defect detection models. This table shows that adding temporal information improves F1 score to detect LOP, LOF, undercut, cold lap, and porosity by 7.95 \%, 6.52 \%, 4.21 \%, 4.21 \%, and 12.5 \%, respectively. Adding the attention technique further improves the F1 score to detect LOP and LOF by 4.21 \% and 1 \%, respectively. The improvement in defect detection performance with attention techniques is due to their ability to focus on the sub-millimeter welding defects, while filtering out irrelevant spatial and temporal information from the larger portions of the image that do not contribute significant insight. Adding the attention module introduces a minimal increase in computational complexity (0.1 GFLOPs due to Softmax and weighted addition), but it improves the F1 score for detecting LOP by 4 \%, making it a worthwhile trade-off. The total GFLOPs for the temporal attention-based model is 0.58.

\begin{table}[H]
    \renewcommand{\arraystretch}{1.3}
    \centering
    \resizebox{0.5\textwidth}{!}{
    \begin{tabular}{lclclcl}
        \hline
        \hline
        \hfil Defect & \hfil Model type & \hfil F1 Score\\
        \hline\hline \\[-3mm]
        \multirow{3}{*}{\hfil LOP} & \hfil Normal & \hfil 0.88  \\
        & \hfil Temporal 3D & \hfil 0.95 \\
        & \hfil \small{Attention Temporal 3D} & \hfil 0.99\\
        \hline
        \multirow{3}{*}{\hfil LOF} & \hfil Normal & \hfil 0.92  \\
        & \hfil Temporal 3D & \hfil 0.98 \\
        & \hfil \small{Attention Temporal 3D} & \hfil 0.99\\
        \hline
        \multirow{3}{*}{\hfil Undercut} & \hfil Normal & \hfil 0.95  \\
        & \hfil Temporal 3D & \hfil 0.99 \\
        & \hfil \small{Attention Temporal 3D} & \hfil 0.99\\
        \hline
        \multirow{3}{*}{\hfil \small{Cold Lap}} & \hfil Normal & \hfil 0.95  \\
        & \hfil Temporal 3D & \hfil 0.99 \\
        & \hfil \small{Attention Temporal 3D} & \hfil 0.99\\
        \hline
        \multirow{3}{*}{\hfil Porosity} & \hfil Normal & \hfil 0.88  \\
        & \hfil Temporal 3D & \hfil 0.99 \\
        & \hfil \small{Attention Temporal 3D} & \hfil 0.99\\
        \hline\hline
    \end{tabular}
    }
    \caption{Performance of the normal, temporal, and attention temporal 3D multi-modal defect detection models}\label{tab:final}
\end{table}

\subsection{Speculative critical information}

\begin{figure}[H]
\centering
\begin{tabular}{p{1.5cm} | p{1.5cm} p{1.5cm} p{1.5cm} p{1.5cm} p{1.5cm}}

Defects & LOP & LOF & Porosity & Undercut & Cold lap \\\\
\hline
Image & 
\includegraphics[width=\linewidth]{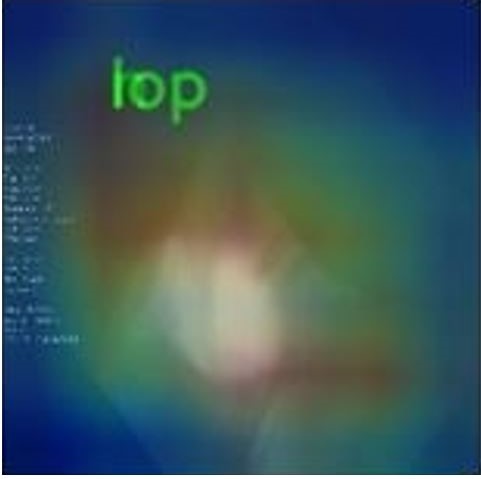} & 
\includegraphics[width=\linewidth]{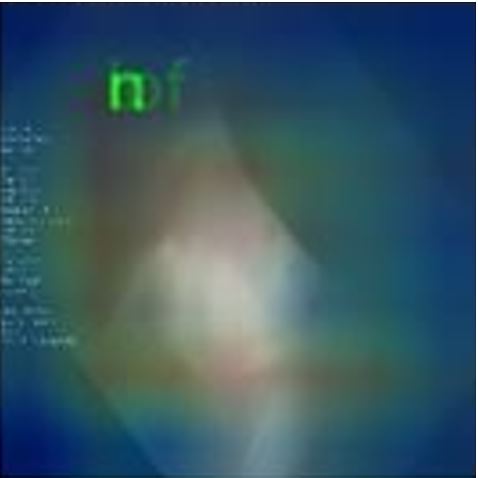} & 
\includegraphics[width=\linewidth]{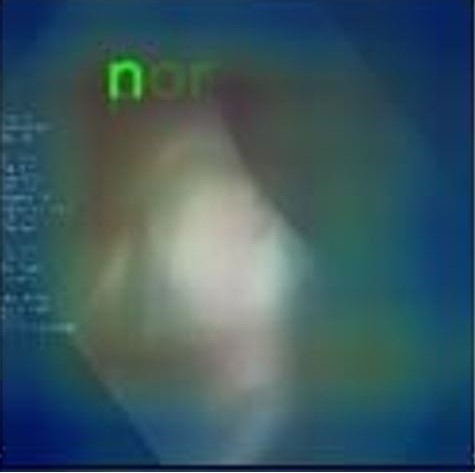} & 
\includegraphics[width=\linewidth]{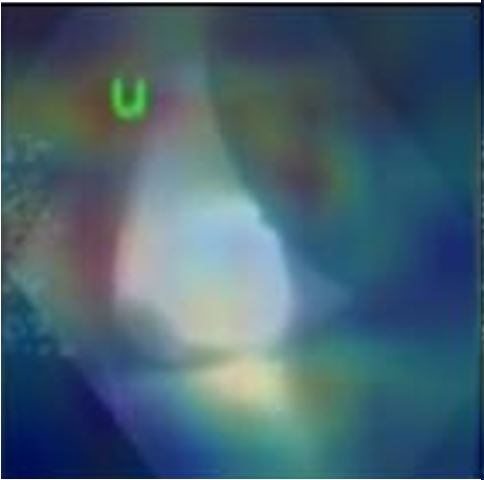} & 
\includegraphics[width=\linewidth]{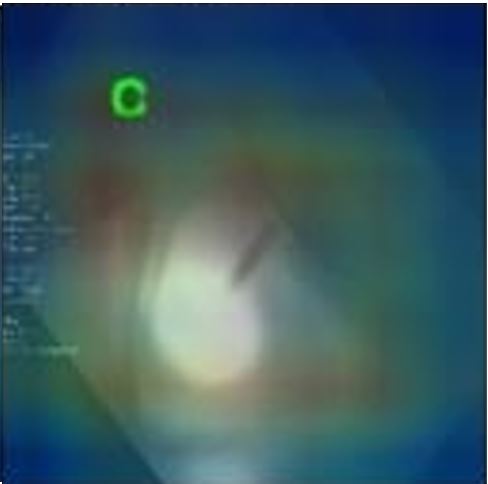} \\\\

Sound & 
\includegraphics[width=\linewidth]{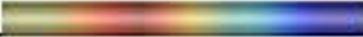} & 
\includegraphics[width=\linewidth]{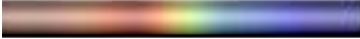} & 
\includegraphics[width=\linewidth]{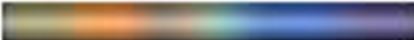} & 
\includegraphics[width=\linewidth]{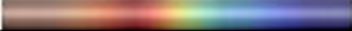} & 
\includegraphics[width=\linewidth]{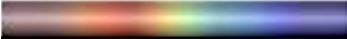} \\\\

\end{tabular}
\caption{Visualization of GradCAM heatmaps on welding images and sound spectrograms for the five target defects. Red pixels show the speculative important areas}
\label{fig:cams}
\end{figure}

For each defect, the proposed critical regions based on the highlighted areas in GardCAM heatmaps are shown in Figure~\ref{fig:cams}. 
The speculative important time-frequency areas in sound spectrograms to detect the five target defects are primarily the first half of the sound spectrogram, with higher activation in earlier time segments.

\begin{figure}[H]
    \centering
    \includegraphics[width=\linewidth]{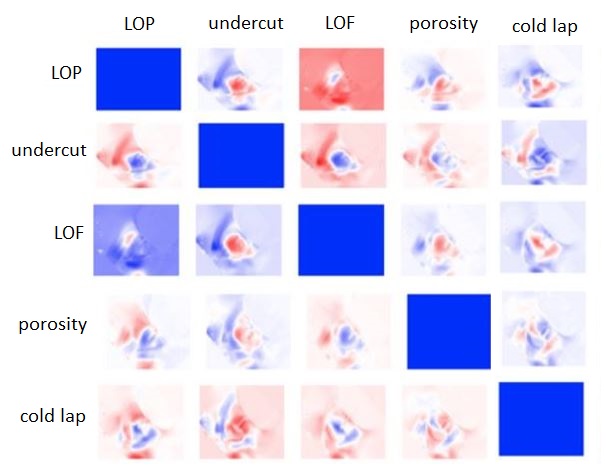}
    \caption{Contrast (pixel-wise difference between the mean of images) between each target defect and other defects. Red pixels show the larger differences and determine the speculative areas that distinguishes each defect from others} 
    \label{fig:contrasts}
\end{figure}

Also, Figure~\ref{fig:contrasts} shows the contrast between each defect and others. This manual feature analysis determines the distinguishing areas to detect each defect. The blue areas to detect LOP, undercut, LOF, porosity, and cold lap are  left leg, right leg, central area, surrounding area, and surrounding areas of molten metal, respectively. These areas are aligned with the results of CAM models in Figure~\ref{fig:cams}.

Table~\ref{tab:XAI} summarizes the speculative critical information in the images to differentiate each target defect from defect-free welding based on GradCAM and manual feature analysis. 

\begin{table}[H]
	\renewcommand{\arraystretch}{1.3}
	\centering
    \centering
	\resizebox{0.7\textwidth}{!}{
		\begin{tabular}{lcl}
                \hline\hline \\[-3mm]
                Defect & Critical information from the image of molten metal\\
			\hline\hline \\[-3mm]
               LOP & left \& right legs \\
                LOF &  central and upper regions\\
               Porosity & surrounding area  \\
               Undercut & right leg\\
               Cold lap & surrounding area \\
			\hline\hline
		\end{tabular}
	}
    \caption{Explaining the speculative critical features in image data by evaluating the results of CAM methods on defect detection models and analyzing the manual features}\label{tab:XAI}
\end{table}

LOP happens when the molten metal does not fully penetrate to the bottom surface of the adjacent solid metal in the weld pool. Based on our results, the critical visual information for LOP is the left and right legs of the molten metal that interacts with the adjacent metal surfaces at the bottom of the molten metal. LOF happens when the molten metal cannot fully fuse to the adjacent metal walls. Our model identifies the central and upper regions of molten metal to detect LOF from welding images. Initiation of porosity can be detected via images. Any impurities that can convert to a gas inside the molten metal can initiate porosity. We proposed the surrounding area of molten metal as the important area to detect porosity initiation from images. Undercut is a drainage-like deep profile of molten metal occurring adjacent to the weld bead due to surface discontinuity. Our model identifies the right leg as the speculative important region in images to detect undercut. Cold lap is the unfused molten metal at the top surface of the molten metal. We proposed upper regions and boundary of molten metal as important visual features to detect cold lap.

In summary, based on the GradCAM explainability analysis, the proposed defect detection models identify key areas for detecting each defect, aligning with expectations from data analysis. The theoretical definitions of each defect further validate the relevance of these areas, confirming that the models focus on the correct regions. This enhances trust in the model's predictions.

\subsection{Preferred modality to detect each defect}

Welding images contain information about the physics of welding that may not be detectable in sound spectrograms. These features including the interaction of buoyancy, marangoni, bernoulli, lorentz, magnet, and gravity forces on the molten metal surface are crucial to detect LOP, LOF, undercut, and cold lap. In Table~\ref{tab:tsne}, we proposed image as the preferred modality to detect these defect based on the t-SNE diagrams of image and sound dataset and the manual image feature analysis. For porosity, we proposed sound as the preferred modality as it is a continuous process that starts at the time of welding and finishes after the solidification is complete. 

\begin{table}[H]
    \renewcommand{\arraystretch}{1.4}  
    \centering
    \small  
    \resizebox{0.75\textwidth}{!}{
        \begin{tabularx}{\textwidth}{>{\centering\arraybackslash}l >{\centering\arraybackslash}c X}  
            \hline\hline
            Defect & Preferred Modality & Analysis \\
            \hline\hline
            LOP & Image & 
                \begin{itemize}
                    \item Image silhouette score (0.41) $>$ sound silhouette score (0.1) (Figure~\ref{fig:tsne_combined}).
                    \item Image-based F1 score (0.85) $>$ sound-based F1 score (0.73) (Table~\ref{tab:final}).
                \end{itemize} \\
            \hline
            LOF & Image & 
                \begin{itemize}
                    \item Image silhouette score (0.18) $>$ sound silhouette score (0.08) (Figure~\ref{fig:tsne_combined}).
                    \item Image-based F1 score (0.58) $>$ sound-based F1 score (0.16) (Table~\ref{tab:final}).
                \end{itemize} \\
            \hline
            Porosity & Sound & 
                \begin{itemize}
                    \item Image silhouette score (0.13) $<$ sound silhouette score (0.16) (Figure~\ref{fig:tsne_combined}).
                    \item Image-based F1 score (0.52) $<$ sound-based F1 score (0.60) (Table~\ref{tab:final}).
                \end{itemize} \\
            \hline
            Undercut & Image & 
                \begin{itemize}
                    \item Image silhouette score (0.64) $>$ sound silhouette score (0.07) (Figure~\ref{fig:tsne_combined}).
                    \item Image-based F1 score (0.73) $>$ sound-based F1 score (0.1) (Table~\ref{tab:final}).
                \end{itemize} \\
            \hline
            Cold Lap & Image & 
                \begin{itemize}
                    \item Image silhouette score (0.48) $>$ sound silhouette score (0.12) (Figure~\ref{fig:tsne_combined}).
                    \item Image-based F1 score (0.85) $>$ sound-based F1 score (0.62) (Table~\ref{tab:final}).
                \end{itemize} \\
            \hline\hline
        \end{tabularx}
    }
    \caption{Determining the preferred modality to detect each target defect by evaluating the t-SNE diagram on image and sound dataset}\label{tab:tsne}
\end{table}

\begin{figure}[H]
    \centering
    \begin{minipage}{0.45\linewidth}
        \centering
        \includegraphics[width=\linewidth]{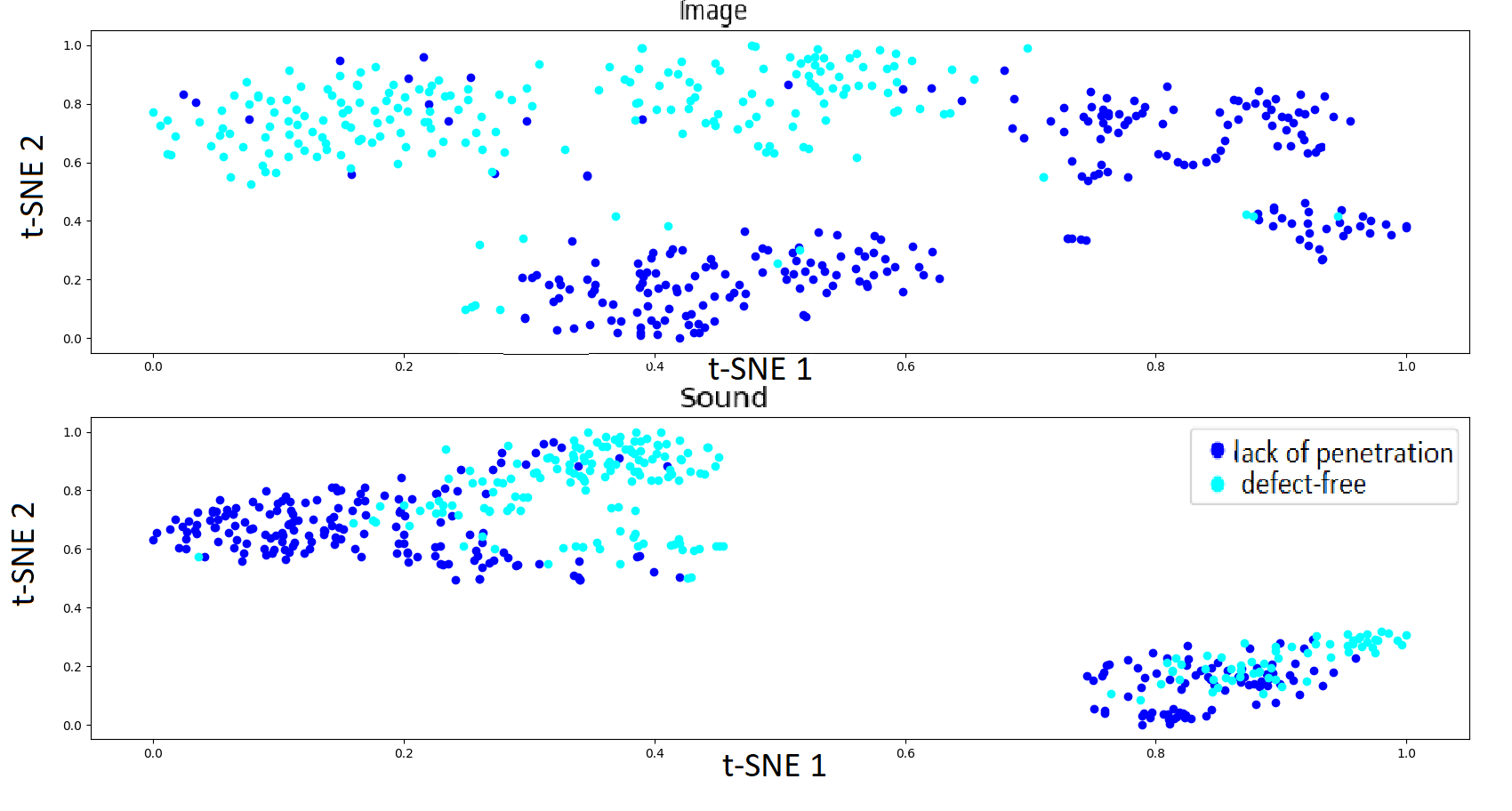}
        \vspace{0.5em}
        \textbf{a)} Lack of Penetration (LOP)
    \end{minipage}
    \hfill
    \begin{minipage}{0.45\linewidth}
        \centering
        \includegraphics[width=\linewidth]{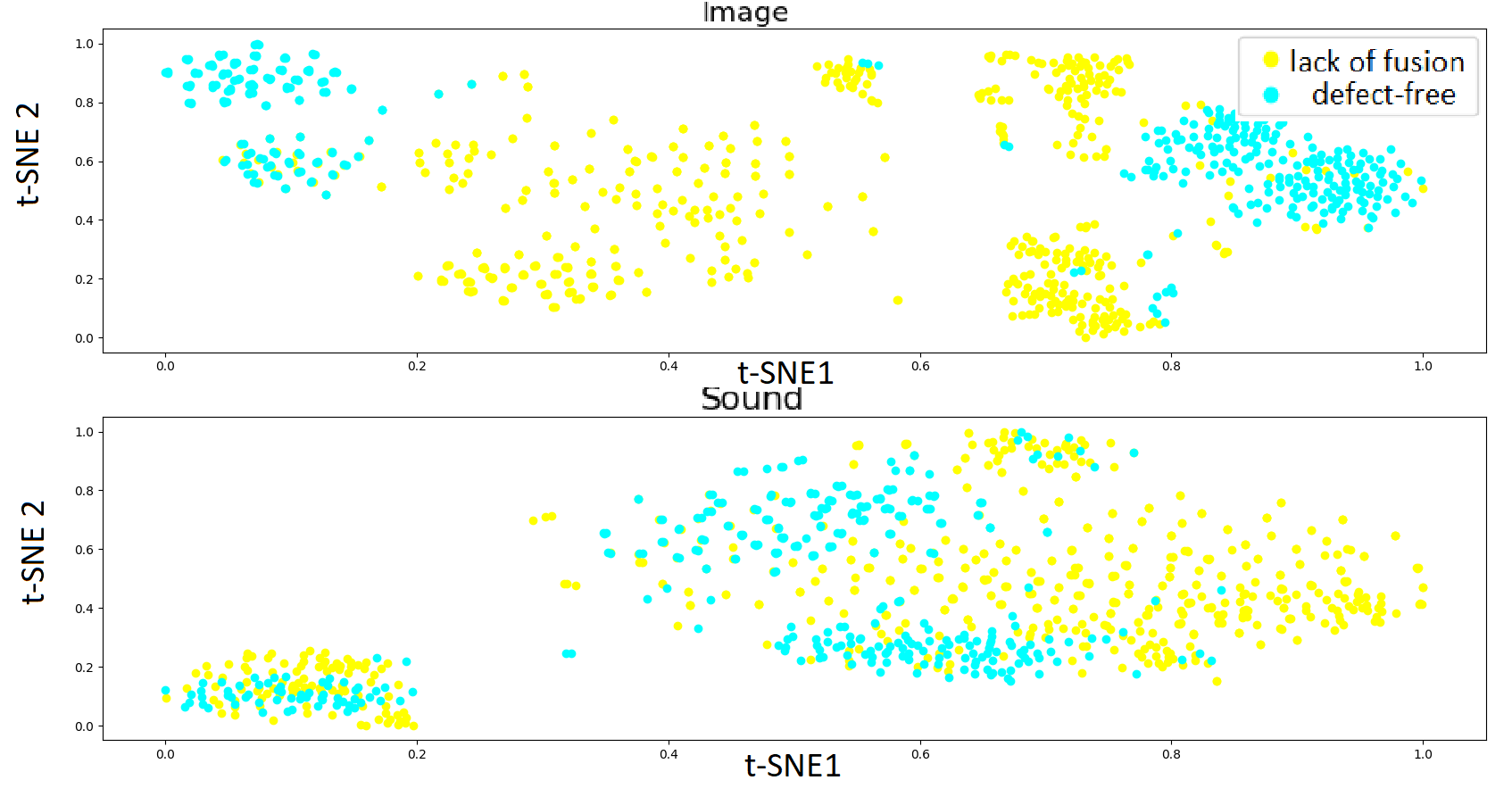}
        \vspace{0.5em}
        \textbf{b)} Lack of Fusion (LOF)
    \end{minipage}

    \vspace{1em} 

    \begin{minipage}{0.45\linewidth}
        \centering
        \includegraphics[width=\linewidth]{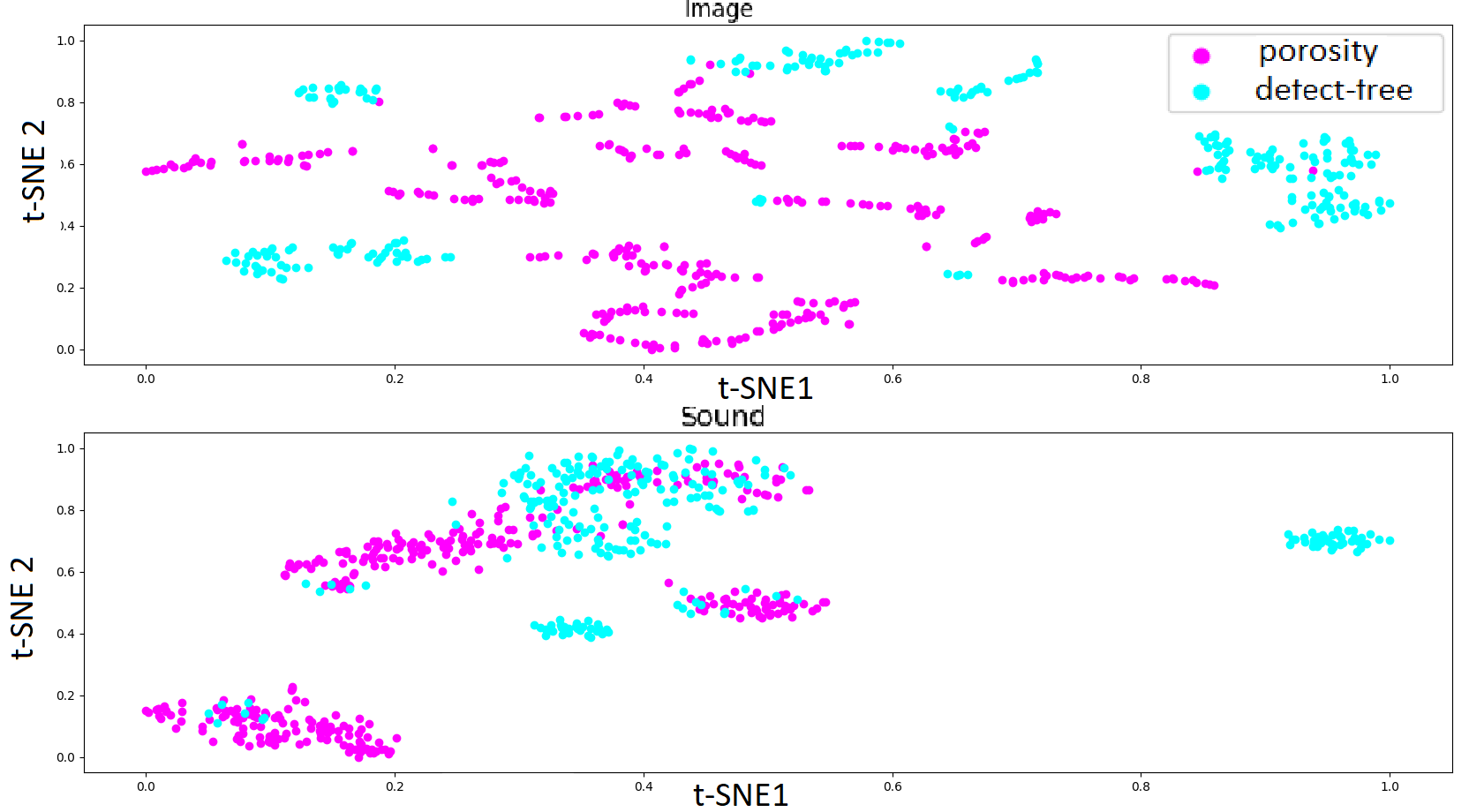}
        \vspace{0.5em}
        \textbf{c)} Porosity
    \end{minipage}
    \hfill
    \begin{minipage}{0.45\linewidth}
        \centering
        \includegraphics[width=\linewidth]{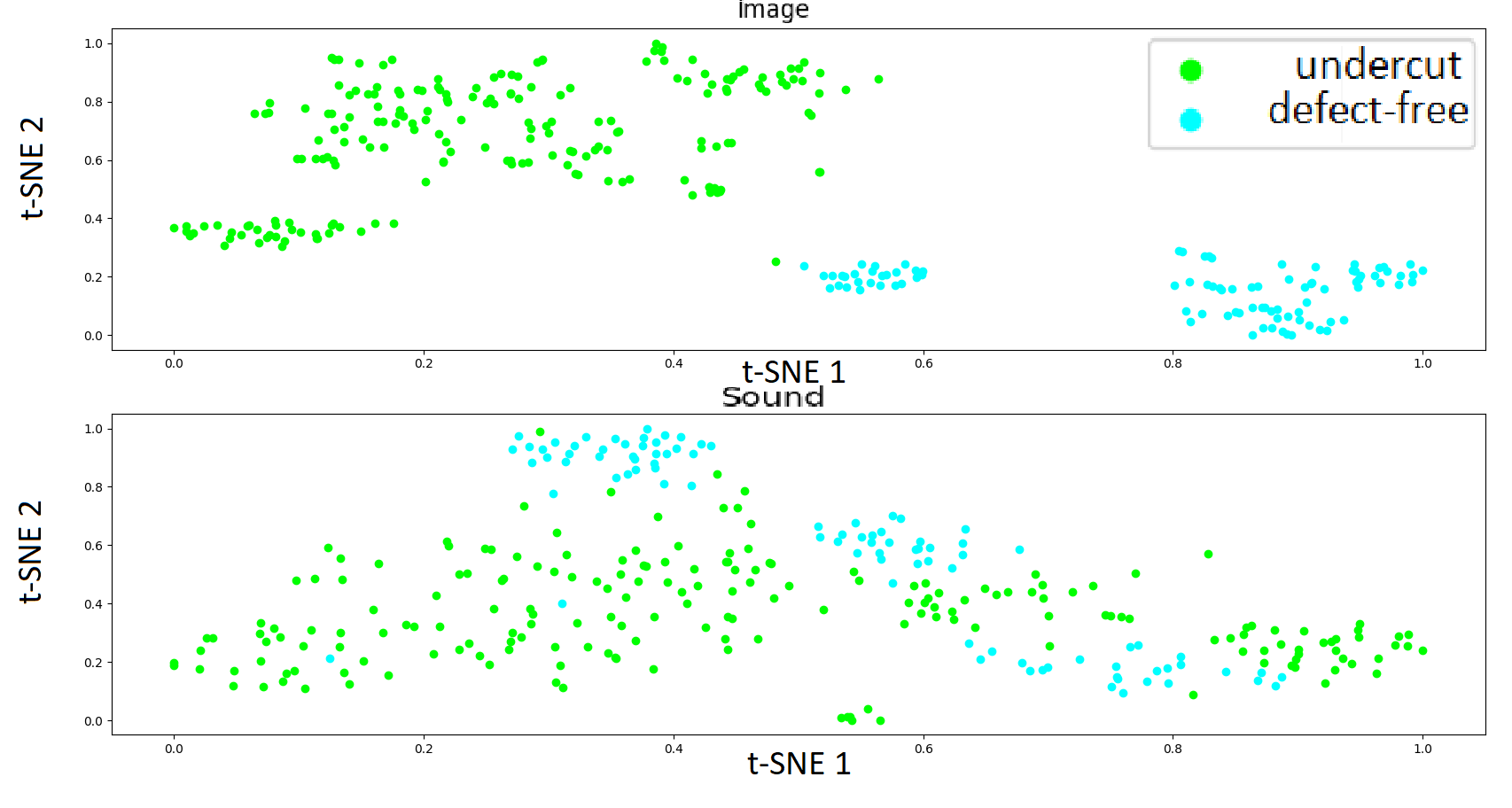}
        \vspace{0.5em}
        \textbf{d)} Undercut
    \end{minipage}

    \vspace{1em} 

    \begin{minipage}{0.45\linewidth}
        \centering
        \includegraphics[width=\linewidth]{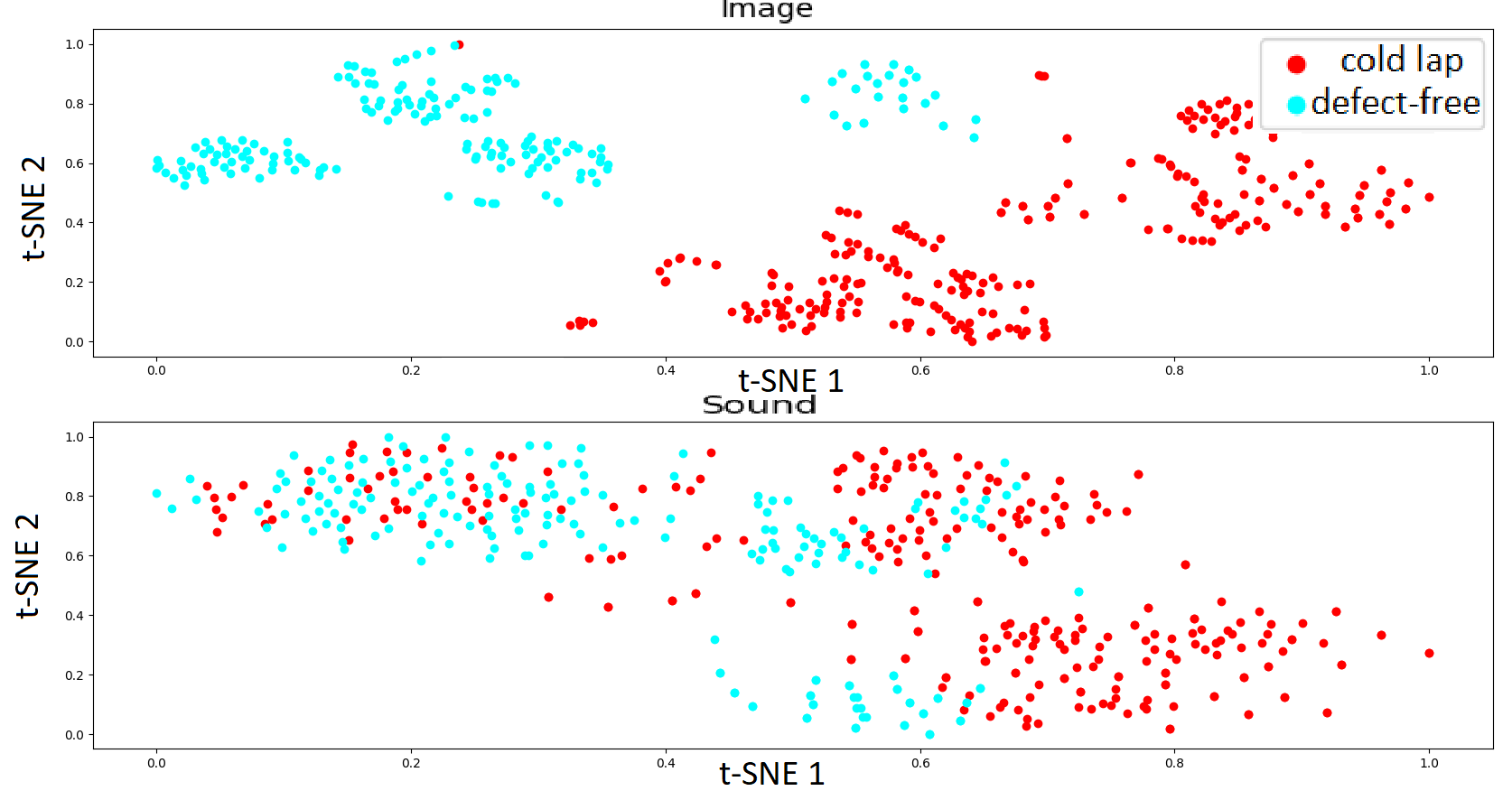}
        \vspace{0.5em}
        \textbf{e)} Cold Lap
    \end{minipage}

    \vspace{1em} 
    \caption{2D t-SNE plots for image and sound modalities showing the distribution of defect-free and defect data for various welding defects. Each point represents a data sample projected into a lower-dimensional space, where the axes correspond to the t-SNE components. Silhouette scores indicate the modality's effectiveness for defect detection}
    \label{fig:tsne_combined}
\end{figure}

\section{Conclusion}
\label{sec:conclusion}

We show that adding the temporal information improves the F1 score by 7.95 \%, 6.52 \%, 4.21 \%, 4.21 \%, and 12.5 \% for LOP, LOF, undercut, cold lap, and porosity detection models in GMAW, respectively compared to the normal multi-modal defect detection. Finally, our results show that adding an attention-based technique~\cite{con2} further improves the performance of the defect detection model by 4.21 \% and 1.02 \% for LOP and LOF, respectively leading to an F1 Score of 0.99 for the five target defects. Table~\ref{tab:sum} summarizes our results. With the explainable AI techniques, we determined the important regions in welding images and sound spectrograms as well as the preferred modality to detect each defect as used in our proposed models. The data for this study is collected in R \& D setting and the defects were intentionally created. In future work, we will evaluate the real-time performance of our proposed defect detection models in production settings. 

\begin{table}[H]
	\renewcommand{\arraystretch}{1.3}
	\centering
    \centering
	\resizebox{0.8\textwidth}{!}{
		\begin{tabular}{lclclclclcl}
                \hline
                \hline
                \hfil Defect &  \hfil Modality & \hfil Normal & \hfil Temporal &  \hfil Attention-based\\
			\hline\hline \\[-3mm]
            \multirow{3}{4em}{\hfil LOP} & \hfil Image & \hfil 0.85 & \hfil 0.91 & \hfil 0.98 \\
            & \hfil Sound & \hfil 0.73 & \hfil 0.73 & \hfil 0.90\\
            & \hfil \small{Multi-modality} & \hfil 0.88 & \hfil 0.95 & \hfil 0.99 \\
            \hline
            \multirow{3}{4em}{\hfil LOF} & \hfil Image & \hfil 0.58 & \hfil 0.96 & \hfil 0.94 \\
            & \hfil Sound & \hfil0.16 & \hfil 0.70 & \hfil 0.85\\
            & \hfil \small{Multi-modality} & \hfil 0.92 & \hfil 0.98 & \hfil 0.99 \\
            \hline
            \multirow{3}{4em}{\hfil \small{Undercut}} & \hfil Image & \hfil 0.73 & \hfil 0.98 & \hfil 0.99 \\
            & \hfil Sound & \hfil 0.10 & \hfil 0.95 & \hfil 0.96\\
            & \hfil \small{Multi-modality} & \hfil 0.95 & \hfil 0.99 & \hfil 0.99 \\
            \hline
            \multirow{3}{4em}{\hfil \small{Cold Lap}} & \hfil Image & \hfil 0.85 & \hfil 0.99 & \hfil 0.99 \\
            & \hfil Sound & \hfil 0.62 & \hfil 0.31 & \hfil 0.87\\
            & \hfil \small{Multi-modality} & \hfil 0.95 & \hfil 0.99 & \hfil 0.99 \\
            \hline
            \multirow{3}{4em}{\hfil Porosity} & \hfil Image & \hfil 0.52 & \hfil 0.95 & \hfil 0.98 \\
            & \hfil Sound & \hfil 0.60 & \hfil 0.74 & \hfil 0.85\\
            & \hfil \small{Multi-modality} & \hfil 0.88 & \hfil 0.99 & \hfil 0.99 \\
			\hline\hline
		\end{tabular}
	}
    \caption{Comparing the F1 Score of the uni-modal and multi-modal defect detection models for the normal, temporal, and attention-based situations}\label{tab:sum}
\end{table}


\section{Acknowledgements}
\label{sec:Acknowledgement}
This work was supported by NSERC under a Canada-Germany 2+2 grant and Mitacs Accelerate scholarship.
We extend our gratitude to Kwang Moo Yi for his consultation on the AI models and the CEO and CTO of Novarc Technologies Inc., Soroush Karimzadeh, and Reza Abdollahi,  for their visionary leadership and support, which made this research possible. We are thankful to Todd Scheerer for his assistance in conducting tests, Ahmad Ashoori for sharing insights on the collaborative robot and its control.

\bibliographystyle{elsarticle-num} 
\bibliography{bib.bib}





\end{document}